\documentclass[10pt,twocolumn,letterpaper]{article}

\usepackage[pagenumbers]{iccv} 

\definecolor{iccvblue}{rgb}{0.21,0.49,0.74}
\usepackage[pagebackref,breaklinks,colorlinks,allcolors=iccvblue]{hyperref}

\def\paperID{1059} 
\def\confName{ICCV}
\def\confYear{2025}

\usepackage{algorithm}
\usepackage{algpseudocode}
\usepackage{multirow}
\usepackage{booktabs}
\usepackage{amsmath}
\usepackage{amssymb}
\usepackage{mathtools}
\usepackage{amsthm}

\usepackage{algpseudocode}

\usepackage{graphicx}
\usepackage{booktabs}  
\usepackage{arydshln}  

\usepackage[capitalize,noabbrev]{cleveref} 

\usepackage{microtype}  
\usepackage{bbm}        
\usepackage{colortbl}   
\usepackage{pifont}     
\definecolor{color3}{gray}{0.95}
\definecolor{color4}{HTML}{00b050}

\newcommand{\ours}{\texttt{AdaSVD}}

\title{$\ours$: Adaptive Singular Value Decomposition for Large Language Models}

\author{
	Zhiteng Li$^{1}$\thanks{Equal contribution}~,\enspace
	Mingyuan Xia$^{1}$\footnotemark[1]~,\enspace
	Jingyuan Zhang$^{1}$,\enspace
	Zheng Hui$^{2}$,\enspace \\
	Haotong Qin$^{3}$,\enspace
	Linghe Kong$^{1}$\footnotemark[2]~,\enspace
	Yulun Zhang$^{1}$\thanks{Corresponding authors: Linghe Kong,  linghe.kong@sjtu.edu.cn, Yulun Zhang, yulun100@gmail.com}~, \enspace
	Xiaokang Yang$^{1}$\\
	\textsuperscript{1}Shanghai Jiao Tong University,\enspace
	\textsuperscript{2}MGTV, Shanhai Academy,\enspace
	\textsuperscript{3}ETH Z\"{u}rich\\
	\vspace{-8mm}
}

\begin{document}
	\maketitle
	
	
	\begin{abstract}
		Large language models (LLMs) have achieved remarkable success in natural language processing (NLP) tasks, yet their substantial memory requirements present significant challenges for deployment on resource-constrained devices. Singular Value Decomposition (SVD) has emerged as a promising compression technique for LLMs, offering considerable reductions in memory overhead. However, existing SVD-based methods often struggle to effectively mitigate the errors introduced by SVD truncation, leading to a noticeable performance gap when compared to the original models. Furthermore, applying a uniform compression ratio across all transformer layers fails to account for the varying importance of different layers. To address these challenges, we propose $\ours$, an adaptive SVD-based LLM compression approach. Specifically, $\ours$ introduces \textbf{adaComp}, which adaptively compensates for SVD truncation errors by alternately updating the singular matrices $\mathcal{U}$ and $\mathcal{V}^\top$. Additionally, $\ours$ introduces \textbf{adaCR}, which adaptively assigns layer-specific compression ratios based on the relative importance of each layer. Extensive experiments across multiple LLM/VLM families and evaluation metrics demonstrate that $\ours$ consistently outperforms state-of-the-art (SOTA) SVD-based methods, achieving superior performance with significantly reduced memory requirements. Code and models of $\ours$ will be available at \url{https://github.com/ZHITENGLI/AdaSVD}.
	\end{abstract}
	
	\setlength{\abovedisplayskip}{2pt}
	\setlength{\belowdisplayskip}{2pt}
	
	\vspace{-3mm}
	\section{Introduction}
	\vspace{-1mm}
	
	\begin{figure}[t]
		\centering
		\includegraphics[width=\linewidth]{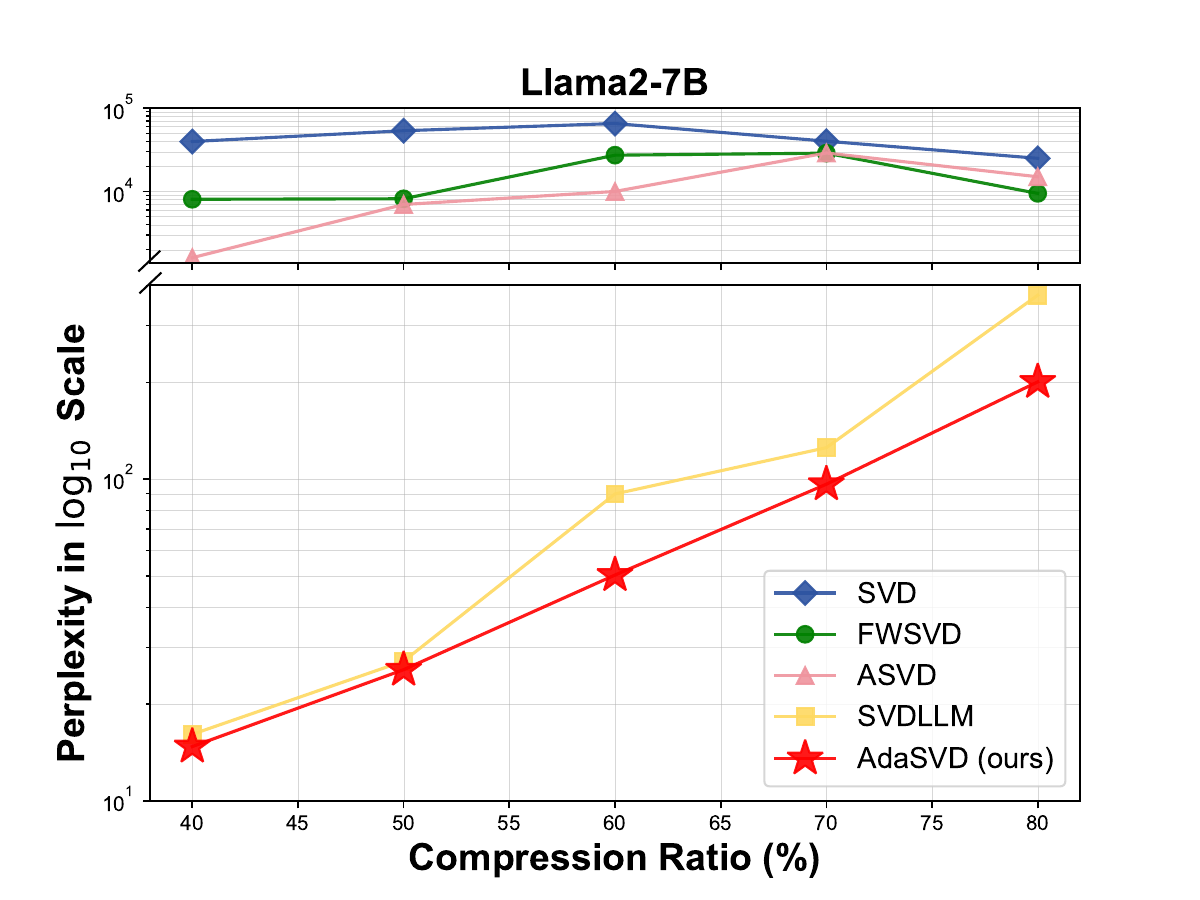} 
		\vspace{-7mm}
		\caption{Comparison between vanilla SVD, FWSVD~\cite{hsu2022fwsvd}, ASVD~\cite{yuan2024asvd}, SVD-LLM~\cite{wang2024svdllm}, and our $\ours$ on WikiText2.}
		\vspace{-7mm}
	\end{figure}
	
	\begin{figure*}[t]
		\centering
		\includegraphics[width=1\textwidth]{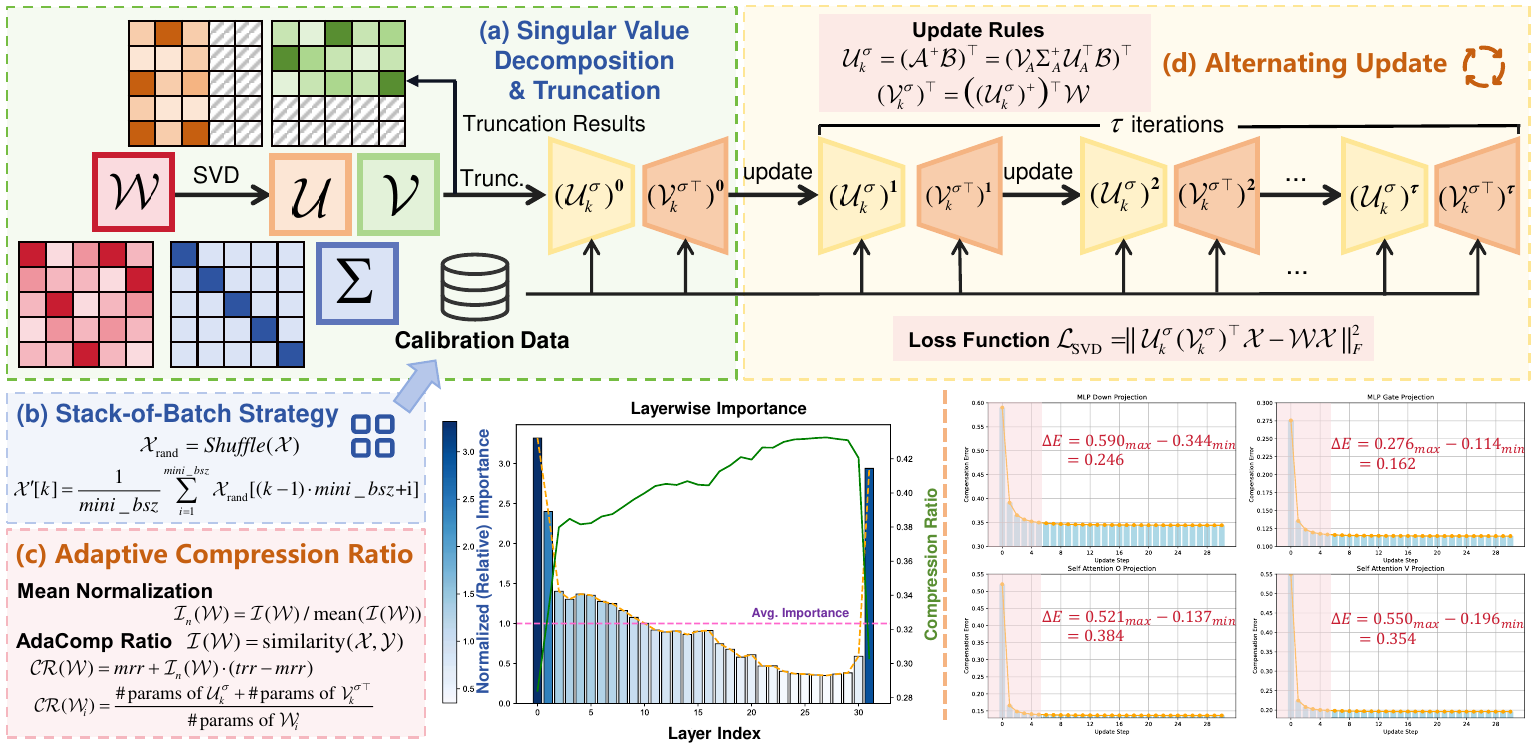}
		\vspace{-7mm}
		\caption{Overview of the proposed $\ours$ method: (a) SVD decomposition and truncation for linear layer weights; (b) Stack-of-batch strategy for efficient use of calibration data under limited GPU memory; (c) Adaptive compression ratio assignment (\textbf{adaCR}) based on layer-wise importance; (d) Adaptive compensation (\textbf{adaComp}) through alternating updates of $\mathcal{U}$ and $\mathcal{V}^\top$.}
		\vspace{-5mm}
		\label{fig:overview}
	\end{figure*}
	
	Recently, large language models (LLMs) based on the Transformer architecture~\cite{vaswani2017attention} have shown remarkable potential across a wide range of natural language processing (NLP) tasks. However, their success is largely driven by their massive scale, with models such as the LLaMA family~\cite{touvron2023llama} and the Open Pre-trained Transformer (OPT) series~\cite{zhang2022opt} containing up to 70B and 66B parameters, respectively. The substantial memory requirements of these models present significant challenges for deploying them on mobile devices. Consequently, the widespread adoption of LLMs remains limited by their immense resource demands~\cite{wan2023efficient, wang2024iot, zhou2024survey}.

	Recent research on large language model (LLM) compression has explored various techniques, including weight quantization~\cite{lin2024awq, frantar2022gptq}, network pruning~\cite{sun2023simple, frantar2023sparsegpt}, low-rank factorization~\cite{wang2024svdllm, zhang2023loraprune, yuan2024asvd}, and knowledge distillation~\cite{zhong2024revisiting, gu2023knowledge}. Among these methods, low-rank factorization using Singular Value Decomposition (SVD)~\cite{hsu2022fwsvd,yuan2024asvd,wang2024svdllm} stands out as a powerful approach for reducing both model size and computational cost. SVD achieves this by decomposing large weight matrices into smaller, low-rank components while preserving model performance. Since LLMs are often memory-bound during inference~\cite{dao2022flashattention,dao2023flashattention}, SVD compression can effectively accelerate model inference by reducing the memory requirements, even when applied solely to the weights. This approach does not require specialized hardware or custom operators, unlike weight quantization, making SVD more versatile across different platforms. Additionally, SVD is orthogonal to other compression techniques~\cite{wang2024svdllm}, allowing it to be combined with methods like weight quantization or network pruning for even greater efficiency, enabling more scalable and adaptable solutions for deploying LLMs.

	Recent advancements in SVD-based LLM compression, including FWSVD~\cite{hsu2022fwsvd}, ASVD~\cite{yuan2024asvd}, and SVD-LLM~\cite{wang2024svdllm}, have significantly improved the low-rank factorization approach, enhancing the overall effectiveness of SVD compression. For example, FWSVD introduces Fisher information to prioritize the importance of parameters, while ASVD accounts for the impact of activation distribution on compression error. SVD-LLM establishes a relationship between singular values and compression loss through the data whitening techniques. While these methods have led to notable improvements in SVD compression, they still face challenges when applied at high compression ratios.

	To bridge the performance gap between compressed and original models at both low and high compression ratios, we revisit SOTA solutions for LLM compression using SVD decomposition. Our analysis highlights two key observations:
	\textbf{First,} low-rank weight compensation after truncating the smallest singular vectors has been largely overlooked or insufficiently explored in prior methods. When truncating parts of the matrices $\mathcal{U}$ and $\mathcal{V}^\top$, the remaining components should be adjusted accordingly to minimize the SVD compression error.
	\textbf{Second,} previous methods typically apply a uniform compression ratio across all transformer layers, failing to account for their varying relative importance. To address this, an importance-aware approach for assigning appropriate compression ratios is necessary.

	To tackle the challenges outlined above, we propose $\ours$, an adaptive SVD-based LLM compression method. \textbf{First,} $\ours$ proposes \textbf{adaComp}, an adaptive compensation technique designed to adjust the weights of $\mathcal{U}$ and $\mathcal{V}^\top$ after SVD truncation. By alternately updating the matrices $\mathcal{U}$ and $\mathcal{V}^\top$, \textbf{adaComp} effectively reduces compression errors in a stable and efficient manner. To optimize the use of calibration data with limited GPU memory, we also introduce a stack-of-batch technique when applying \textbf{adaComp}.
	\textbf{Second,} $\ours$ proposes \textbf{adaCR}, a method that assigns adaptive compression ratios to different layers based on their importance. With the target compression ratio fixed, this strategy significantly improves performance compared to using a uniform compression ratio across all layers.

	Our key contributions are summarized as follows:
	\begin{itemize}
		\item We propose \textbf{adaComp}, a novel adaptive compensation method for SVD truncation. By alternately updating $\mathcal{U}$ and $\mathcal{V}^\top$ and employing the stack-of-batch technique, we effectively and stably minimize compression error.
		
		\item We propose \textbf{adaCR}, an adaptive compression ratio method that assigns layer-specific compression ratios according to their relative importance in LLMs. This importance-aware approach outperforms the previously used uniform compression ratio method.
		
		\item Extensive experiments on LLMs/VLMs demonstrate that our method, $\ours$, significantly outperforms the previous SOTA SVD-based LLM compression method, SVD-LLM, effectively narrowing the performance gap between compressed and original models.

	\end{itemize}

	\begin{algorithm*}[t]
		\caption{Pseudocode of \ours}
		\begin{algorithmic}[1] 
			\State \textbf{Inputs:} LLM $\mathcal M$, Calib Data $\mathcal C$, Bucket Size $M$, Target Retention Ratio $trr$, Min Retention Ratio $mrr$, Update Iteration $k$
			\State \textbf{Outputs:} Updated Model $\mathcal M'$ by \ours
			\Procedure{\ours}{$\mathcal{M, C}, trr, mrr, k$} 
			\State $\mathcal{X} \gets$ \textproc{Get\_calib($\mathcal{C}$)}  \Comment{Randomly collect samples as calibration data}
			
			\State $\mathcal{X}'[1],\mathcal{X}'[2],...,\mathcal{X}'[M] \gets $ \textproc{SOB($\mathcal{X}, M$)} \Comment{Shuffle samples and utilize stack-of-batch (SOB) strategy}
			
			\State $\text{Set}_\mathcal{S} \gets \textproc{Whitening}(\mathcal M, \mathcal{X}')$, $\text{Set}_\mathcal{SVD} \gets \emptyset$, $\text{Set}_\mathcal{W} \gets \mathcal{M}$ \Comment{Initialize sets of decomposed matrices and weights}
			
			\State $\text{Set}_\mathcal{CR} \gets$ \textproc{Layer\_CR($\mathcal{M}, \mathcal{X}', trr, mrr$)}
			\Comment{Calculate layerwise importance and compression ratio}
			
			\For{layer $i$ \textbf{in} language model $\mathcal{M}$}
			\State $\mathcal{W}_i \gets \text{Set}_\mathcal{W}(i)$, $\mathcal{S}_i \gets \text{Set}_\mathcal{S}(\mathcal{W}_i)$  \Comment{Extract the whitening matrix of current weight $\mathcal{W}_i$} 
			
			\State $\mathcal{U}_i, \Sigma_i, \mathcal{V}_i \gets$ \textproc{SVD($\mathcal{W}_i\mathcal{S}_i$)}
			\Comment{Apply Singular Value Decomposition}
			
			\State $\Sigma' \gets$ \textproc{Trunc($\Sigma_i$)}, ($\mathcal{U}_i', \mathcal{V}_i') \gets$ \textproc{Trunc\_UV($\mathcal{U,V},\Sigma'$)}
			\Comment{Apply adaptive compression ratio and truncation}
			
			\State $\text{Set}_{\mathcal{SVD}} \gets (\mathcal{U}_i', \mathcal{V}_i') \cup \text{Set}_{\mathcal{SVD}}$ 
			
			\EndFor
			\State $\mathcal{M}' \gets$ \textproc{Ada\_Update$(\mathcal{M, \mathcal{X}', \text{Set}_{SVD}}, k)$}
			\Comment{Utilize alternate update for $\mathcal{U}_i', \mathcal{V}_i'$ with iteration $k$}
			
			\State \Return{$\mathcal{M'}$}
			\EndProcedure
		\end{algorithmic}
		\label{algo:framework}
	\end{algorithm*}

	\vspace{-2mm}
	\section{Related Works}
	\vspace{-2mm}
	\subsection{LLM Compression Techniques}
	\vspace{-2mm}
	Recent advancements in model compression techniques have significantly enhanced the efficiency of deploying LLMs while maintaining their performance. Widely explored approaches include weight quantization~\cite{frantar2022gptq, lin2024awq}, network pruning~\cite{frantar2023sparsegpt, ma2023llmpruner, yang2024laco, gromov2024unreasonable, ashkboos2024slicegpt}, and hybrid methods~\cite{dong2024stbllm}.
	In unstructured pruning, SparseGPT~\cite{frantar2023sparsegpt} prunes weights based on their importance, as determined by the Hessian matrix. However, it faces challenges in achieving optimal speedup, particularly due to hardware compatibility issues. Structured pruning methods, in contrast, are more hardware-friendly. LLM-Pruner~\cite{ma2023llmpruner} selectively removes non-critical coupled structures using gradient information. LaCo~\cite{yang2024laco} introduces a layer-wise pruning strategy, where subsequent layers collapse into preceding ones. ~\citet{gromov2024unreasonable} explores the effectiveness of basic layer-pruning techniques combined with parameter-efficient fine-tuning (PEFT). Additionally, SliceGPT~\cite{ashkboos2024slicegpt} has pioneered post-training sparsification, emphasizing the importance of layer removal order for optimal performance.
	Quantization techniques offer another significant avenue for compression. GPTQ~\cite{frantar2022gptq} applies layer-wise quantization and reduces quantization errors through second-order error compensation. AWQ~\cite{lin2024awq} introduces activation-aware weight quantization, employing a scale transformation between weights and activations. Moreover, BiLLM~\cite{huang2024billm} and ARB-LLM~\cite{li2024arb} achieve further compression to 1-bit while maintaining remarkable performance. More recently, STB-LLM~\cite{dong2024stbllm} combines 1-bit quantization with pruning to achieve even greater memory reduction for LLMs.
	However, many of these compression techniques face challenges related to hardware compatibility, often requiring custom CUDA kernels~\cite{dong2024stbllm} to enable real-time inference speedup.

	\vspace{-1.2mm}
	\subsection{SVD-based LLM Compression}
	\vspace{-1.2mm}
	Singular Value Decomposition (SVD) is a widely used technique for reducing matrix size by approximating a matrix with two smaller, low-rank matrices~\cite{GOLUB1987317}. Although SVD-based methods have demonstrated potential in compressing LLMs, their full capabilities remain underexplored. Standard SVD typically focuses on compressing the original weight matrix without considering the significance of individual parameters, which can lead to considerable compression errors. To address this, \citet{hsu2022languagemodel} introduced FWSVD, which incorporates Fisher information to weight the importance of parameters. However, this method requires complex gradient calculations, making it resource-intensive. Another limitation of standard SVD is the impact of activation distribution on compression errors. To mitigate this, \citet{yuan2024asvd} proposed ASVD, which scales the weight matrix with a diagonal matrix that accounts for the influence of input channels on the weights. Subsequently, \citet{wang2024svdllm} introduced SVD-LLM, which establishes a connection between singular values and compression loss. This work demonstrates that truncating the smallest singular values after data whitening effectively minimizes compression loss. Despite these advancements, existing methods still exhibit significant accuracy loss at higher compression ratios and lack a comprehensive approach for compensating compressed weights after SVD truncation. Furthermore, most methods apply a uniform compression ratio across all transformer layers, overlooking the varying importance of different layers. $\ours$ seeks to address these limitations by proposing an adaptive compensation method (\textbf{adaComp}) and an importance-aware adaptive compression ratio method (\textbf{adaCR}).

	\vspace{-2.5mm}
	\section{Method}
	\vspace{-2.5mm}
	\textbf{Overview.\quad} As illustrated in~\cref{fig:overview}, our $\ours$ integrates adaptive compensation for SVD truncation (\textbf{adaComp}) with an adaptive importance-aware compression ratio method (\textbf{adaCR}). In~\cref{sec:adacom}, we first describe how \textbf{adaComp} compensates for SVD truncation. Next, in~\cref{sec:adacr}, we detail how \textbf{adaCR} determines the compression ratio based on layer importance. The pseudocode of $\ours$ is shown in~\cref{algo:framework}, and pseudocodes for \textbf{adaComp} and \textbf{adaCR} are provided in the supplementary file.
	
	\vspace{-2mm}
	\subsection{Adaptive Compensation for SVD Truncation}
	\label{sec:adacom}
	\vspace{-1.5mm}
	SVD compression first applies SVD decomposition for matrix $\mathcal{W}$, and then truncates the smallest singular values:
	\begin{align}
		\mathcal{W} = \mathcal{U}\Sigma \mathcal{V}^\top \approx \mathcal{U}_k\Sigma_k\mathcal{V}_k^\top = \widehat{\mathcal{W}},
	\end{align}
	where $\Sigma_k$ indicates the retaining top-k largest singular values, $\mathcal{U}_k$ and $\mathcal{V}_k^\top$ represent the corresponding retaining singular vectors. Moreover, the diagonal matrix $\Sigma_k$ can be further absorbed into $\mathcal{U}_k$ and $\mathcal{V}_k^\top$ by
	\begin{align}
		\mathcal{U}_k^\sigma &= \mathcal{U}_k\Sigma_k^\frac{1}{2}, \ \mathcal{V}_k^\sigma = \mathcal{V}_k\Sigma_k^\frac{1}{2},\\ \widehat{\mathcal{W}}&=\mathcal{U}_k\Sigma_k\mathcal{V}_k^\top = \mathcal{U}_k^\sigma(\mathcal{V}_k^\sigma)^\top.
	\end{align}
	The truncation of the smallest singular values minimizes the compression error with respect to $\mathcal{W}$, ensuring that $||\mathcal{U}_k^\sigma(\mathcal{V}_k^\sigma)^\top-\mathcal{W}||_F^2$ is minimized, which we refer to as the vanilla SVD method. However, this approach does not fully account for the practical effects of $\mathcal{X}$. To address this limitation, we introduce a more application-relevant metric for the SVD compression error, defined as follows:
	\begin{align}
		\mathcal{L}_\text{SVD}&=||\widehat{\mathcal{W}}\mathcal{X}-\mathcal{WX}||_F^2 \notag\\
		&=||\mathcal{U}_k^\sigma(\mathcal{V}_k^\sigma)^\top \mathcal{X}-\mathcal{WX}||_F^2.
	\end{align}
	Previous works~\cite{hsu2022languagemodel, yuan2024asvd, wang2024svdllm} have made significant efforts to minimize $\mathcal{L}_\text{SVD}$. However, some of them involve complex and time-consuming preprocessing steps. Furthermore, they still face substantial challenges in effectively mitigating the large errors that arise under high compression ratios, particularly when truncating 60\% or more of the parameters.
	
	\begin{figure}[t]
		\centering
		\includegraphics[width=1\linewidth]{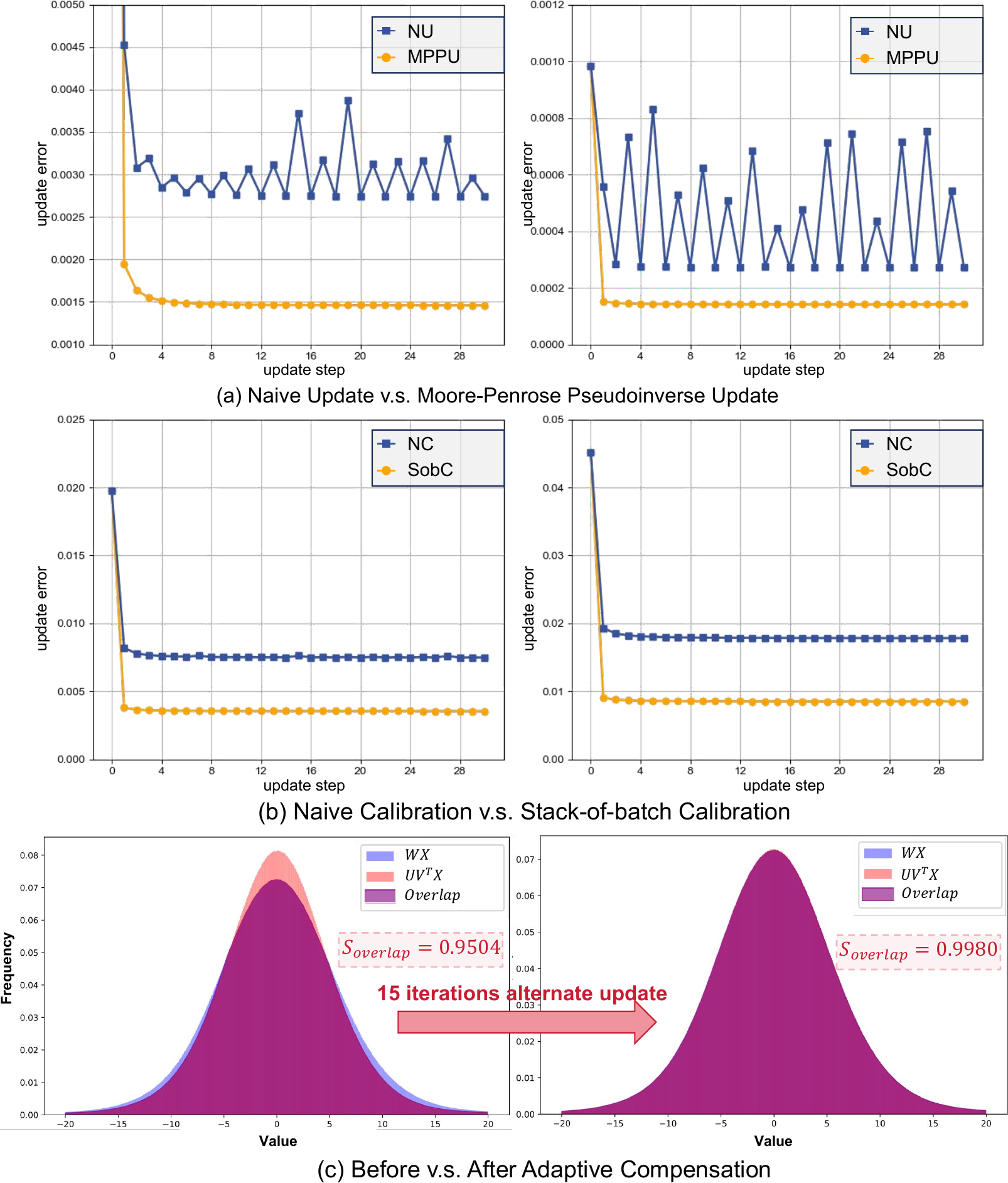}
		\vspace{-7mm}
		\caption{Adaptive compensation for SVD truncation (\textbf{adaComp}). (a) Comparison between naive (NU) and Moore-Penrose pseudoinverse update (MPPU). (b) Comparison between naive (NC) and stack-of-batch calibration strategy (SobC). (c) Distribution comparison before and after applying \textbf{adaComp}. }
		\vspace{-6mm}
		\label{fig:3.1}
	\end{figure}
	
	To compensate for the error attributed to SVD truncation, we need to optimize the following objective:
	\begin{align}
		\mathcal{U}_k^\sigma, {\mathcal{V}_k^\sigma}^\top &= \arg\min_{\mathcal{U}_k^\sigma, {\mathcal{V}_k^\sigma}^\top} \| \mathcal{U}_k^\sigma {\mathcal{V}_k^\sigma}^\top \mathcal{X} - \mathcal{WX} \|_F^2.
	\end{align}
	A straightforward approach is to compute the partial derivatives of the SVD compression objective with respect to $\mathcal{U}_k^\sigma$ and ${\mathcal{V}_k^\sigma}^\top$, resulting in the following expressions (additional details can be found in the supplementary file):
	\begin{align}
		&\frac{\partial \mathcal{L}_\text{SVD}}{\partial \mathcal{U}_k^\sigma} = 0 \notag\\
		&\quad\Rightarrow \mathcal{U}_k^\sigma = \mathcal{WX} \mathcal{X}^\top \mathcal{V}_k^\sigma((\mathcal{V}_k^\sigma)^\top \mathcal{X} \mathcal{X}^\top \mathcal{V}_k^\sigma)^{-1}, \\
		&\frac{\partial \mathcal{L}_\text{SVD}}{\partial {\mathcal{V}_k^\sigma}^\top} = 0 \notag\\
		&\quad\Rightarrow {\mathcal{V}_k^\sigma}^\top = ((\mathcal{U}_k^\sigma)^\top \mathcal{U}_k^\sigma)^{-1}(\mathcal{U}_k^\sigma)^\top \mathcal{W}.
	\end{align}
	However, this method involves computing the matrix inverse, which can lead to unstable updates and significant compression errors, as shown in~\cref{fig:3.1} (a). To mitigate the issue of numerical instability, we propose a two-fold strategy to enhance the update quality of $\mathcal{U}_k^\sigma$ and ${\mathcal{V}_k^\sigma}^\top$.
	
	\begin{figure*}[htbp]
		\centering
		\includegraphics[width=1\textwidth]{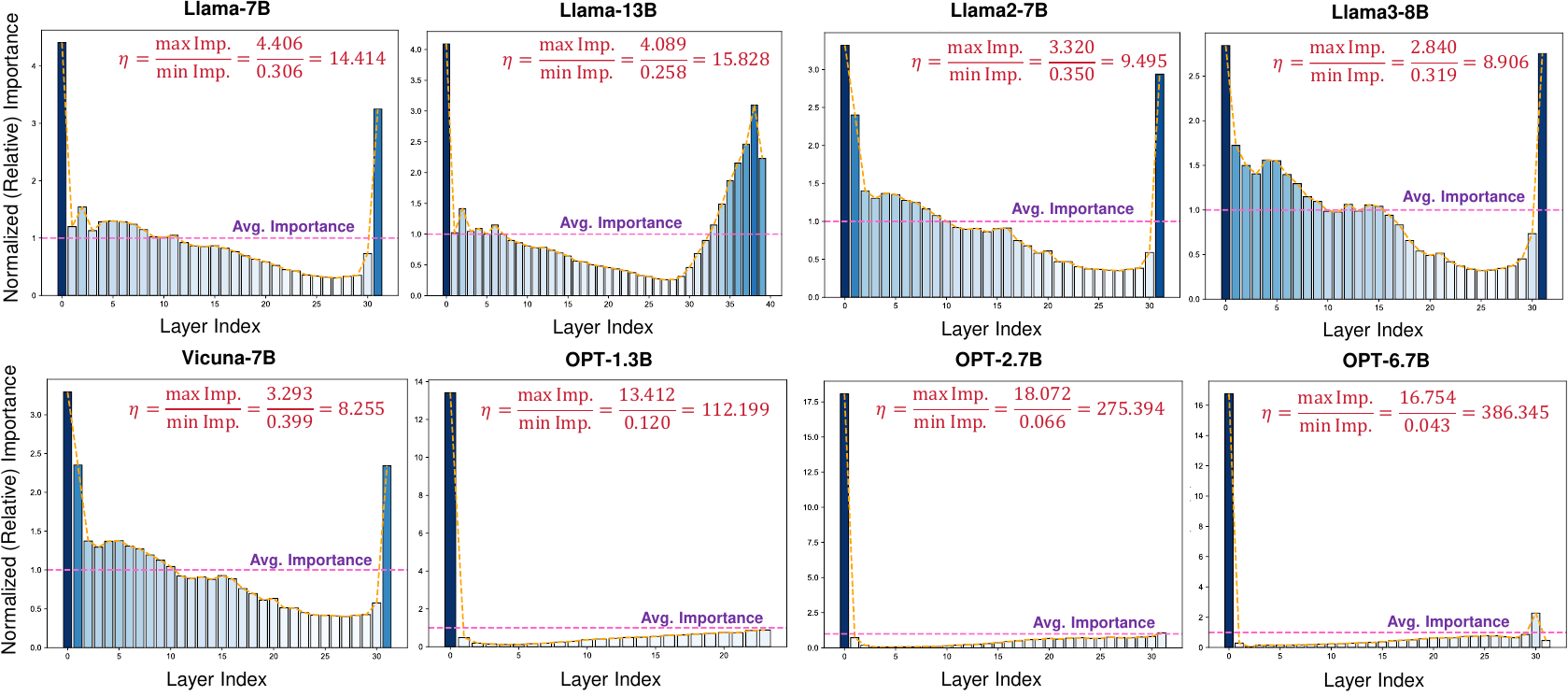}
		\vspace{-7mm}
		\caption{Layer-wise relative importance of different LLMs. The importance across different layers varies significantly, and the first layer always weight most importance. More layer-wise importance visualization can be found in the supplementary file.}
		\vspace{-3.5mm}
		\label{fig:layer_importance}
	\end{figure*}
	
	\textbf{First}, the optimization objective for $\mathcal{U}_k^\sigma$ is reformulated as a Least Squares Estimation (LSE) problem, where ${\mathcal{V}_k^\sigma}^\top \mathcal{X}$ is treated as the input and $\mathcal{WX}$ as the output:
	\begin{align}
		\mathcal{U}_k^\sigma &= \arg\min_{\mathcal{U}_k^\sigma} \| \mathcal{A}(\mathcal{U}_k^\sigma)^\top  - \mathcal{B} \|_F^2,
	\end{align}
	where $\mathcal{A}=\mathcal{X}^\top \mathcal{V}_k^\sigma$ and $\mathcal{B}=(\mathcal{WX})^\top$. Since 
	$\mathcal{A}$ is typically not a square matrix and may not be full rank, we first apply SVD to 
	$\mathcal{A}$ to enhance numerical stability:
	\begin{align}
		\mathcal{A} = \mathcal{U}_\mathcal{A}\Sigma_\mathcal{A}\mathcal{V}_\mathcal{A}^\top,
	\end{align}
	and then obtain the solution for $\mathcal{U}_k^\sigma$ by using the Moore-Penrose pseudoinverse~\cite{penrose1955generalized} of $\mathcal{A}$:
	\begin{align}
		\mathcal{U}_k^\sigma = (\mathcal{A}^+\mathcal{B})^\top = (\mathcal{V}_\mathcal{A}\Sigma_\mathcal{A}^+\mathcal{U}_\mathcal{A}^\top \mathcal{B})^\top,
	\end{align}
	where $\Sigma_A^+$ denotes the Moore-Penrose pseudoinverse of $\Sigma_A$:
	\begin{align}
		\Sigma_A &= \text{diag}(\sigma_1, \sigma_2, \dots, \sigma_n), \\
		\Sigma_A^+ &= \text{diag} \left( \sigma_1^{-1} \mathbbm{1}_{\sigma_1 \neq 0}, \sigma_2^{-1} \mathbbm{1}_{\sigma_2 \neq 0}, \dots, \sigma_n^{-1} \mathbbm{1}_{\sigma_n \neq 0} \right).
	\end{align}
	Similarly, we update ${\mathcal{V}_k^\sigma}^\top$ using the Moore-Penrose pseudoinverse of $\mathcal{U}_k^\sigma$ to handle numerical instability:
	\begin{align}
		{\mathcal{V}_k^\sigma}^\top &= \arg\min_{{\mathcal{V}_k^\sigma}^\top} \| \mathcal{U}_k^\sigma {\mathcal{V}_k^\sigma}^\top \mathcal{X} - \mathcal{WX} \|_F^2 \notag\\
		&= {\Big((\mathcal{U}_k^\sigma)^+\Big)}^\top \mathcal{W}.
	\end{align}
	As shown in~\cref{fig:3.1} (a), by reformulating the optimization objective as an LSE problem and solving for $\mathcal{U}$ and $\mathcal{V}^\top$ using the Moore-Penrose pseudoinverse, we achieve a smooth curve that consistently reduces compression error stably.
	
	\textbf{Second}, since the update rule incorporates the calibration data $\mathcal{X}$, ideally, a large volume of $\mathcal{X}$ would yield better results. However, during our experiments, we found that extending $\mathcal{X}$ to just 32 samples on an 80GB GPU is challenging. To address this, we propose a \textbf{stack-of-batch} strategy that enables the utilization of more calibration data without increasing memory overhead. Specifically, given $N$ calibration samples and a bucket size $M$ (the maximum number of samples that can fit within the fixed GPU memory), we randomly sample $mini\_bsz=\lceil\frac{N}{M}\rceil$ samples into one bucket by taking their mean value as follows:
	\begin{align}
		\mathcal{X}_{\text{rand}} &= \textit{Shuffle}(\mathcal{X}), \\
		\mathcal{X}'[k] &= \frac{1}{mini\_bsz}\sum_{i=1}^{mini\_bsz} \mathcal{X}_{\text{rand}}[(k-1) \cdot mini\_bsz + i],
	\end{align}
	where $k = 1, 2, \dots, M$, and cardinality $|\mathcal{X}'|=M$.
	As shown in~\cref{fig:3.1} (b), integrating the \textbf{stack-of-batch} strategy further reduces the compression error.

	As shown in~\cref{fig:overview}, 
	to compensate for the error attributed to SVD truncation, we propose an adaptive method to subsequently update $\mathcal{U}_k^\sigma$ and $\mathcal{V}_k^\sigma$ with the above update rules.
	Moreover, the adaptation of $\mathcal{U}_k^\sigma$ and $\mathcal{V}_k^\sigma$ can be alternatively applied until convergence, where the update sequence over $\tau$ iterations can be expressed as
	\begin{align}
		\boxed{(\mathcal{U}_k^\sigma)^\mathbf{1}
			\rightarrow ({\mathcal{V}_k^\sigma}^\top)^\mathbf{1}} &\rightarrow \boxed{(\mathcal{U}_k^\sigma)^\mathbf{2} \rightarrow ({\mathcal{V}_k^\sigma}^\top)^\mathbf{2}} \notag\\
		\rightarrow \cdots &\rightarrow \boxed{(\mathcal{U}_k^\sigma)^{\boldsymbol{\tau}} \rightarrow ({\mathcal{V}_k^\sigma}^\top)^{\boldsymbol{\tau}}},
	\end{align}
	where $(\mathcal{U}_k^\sigma)^{\boldsymbol{\tau}}$ and $({\mathcal{V}_k^\sigma}^\top)^{\boldsymbol{\tau}}$ denote the updated singular matrices after $\tau$-th iteration, respectively, while the region bounded by $\boxed{\phantom{0000}}$ corresponding to one iteration of alternative update.
	As shown in~\cref{fig:3.1} (c), the gap between the outputs of the compressed and original models narrows after alternative updates. The overlapping area rapidly increases after just a few iterations. More visual comparisons are shown in the supplementary file.
	
	Notably, our adaptive compensation can be integrated with data whitening proposed by~\citet{wang2024svdllm} and~\citet{liu2024eora}, further reducing the SVD truncation error.
	
	\begin{table*}[t]
		\centering

		\resizebox{1\textwidth}{!}{
		\begin{tabular}{c|c|c|c|c|c|c|c|c|c|c
			}
			\toprule
			{\textsc{Ratio}}       & {\textsc{Method}}    & ~~~~WikiText-2{$\downarrow$}~~~~ & PTB{$\downarrow$} & ~~{C4{$\downarrow$}}~~ & ~~Mmlu~~ & ~ARC\_e~ & ~WinoG.~ & ~HellaS.~  & ~~PIQA~~ & ~\textbf{Average{$\uparrow$}}~        \\ \midrule
			{\color[HTML]{9B9B9B}0\%}  & {\color[HTML]{9B9B9B}Original}  & {\color[HTML]{9B9B9B}5.68}   & {\color[HTML]{9B9B9B}8.35}     & {\color[HTML]{9B9B9B}7.34}      & {\color[HTML]{9B9B9B} 45.30} & {\color[HTML]{9B9B9B} 74.62} & {\color[HTML]{9B9B9B} 69.22} & {\color[HTML]{9B9B9B} 76.00}  & {\color[HTML]{9B9B9B} 79.11} & {\color[HTML]{9B9B9B} 68.85}       \\ \midrule
			{\multirow{5}{*}{40\%}}   & {SVD}      & 39,661.03           & 69,493.00          & {56,954.00} & \textbf{26.51} &  26.39 & 48.62  &  25.64 & 52.99    & 36.03 \\ 
			
			{} & {FWSVD~\cite{hsu2022languagemodel}}      & 8,060.35           & 9,684.10          & 7,955.21              & 25.74  &26.05   &50.20   &25.70      &52.39     &36.01  \\
			
			{} & {ASVD~\cite{yuan2024asvd}}      & 1,609.32           & 7,319.49          & 1,271.85              &24.35   &26.81   &49.49   &25.83     &53.81     &36.06  \\
			
			{} & {SVD-LLM~\cite{wang2024svdllm}}      &16.11  &719.44          & {61.95}              & 22.97  & 36.99  & 56.04  & 30.49   & 56.96  & 40.69 \\
			\cmidrule{2-11} 
			{}                     & \cellcolor{purple!10}{\textbf{$\ours$}}  & \cellcolor{purple!10}\textbf{14.76} ~\footnotesize($\downarrow$8\%)  & \cellcolor{purple!10}\textbf{304.62}~\footnotesize($\downarrow$58\%) & \cellcolor{purple!10}{\textbf{56.98}}~\footnotesize($\downarrow$8\%)   & \cellcolor{purple!10}23.63               & \cellcolor{purple!10}\textbf{41.12}               & \cellcolor{purple!10}\textbf{58.17}               & \cellcolor{purple!10}\textbf{31.75}                  & \cellcolor{purple!10}\textbf{58.49}               
			& \cellcolor{purple!10}\textbf{42.63}    \\

			\midrule
			{\multirow{5}{*}{50\%}}                     & {SVD}      & 53,999.48           & 39,207.00          & {58,558.00} & \textbf{25.43}  & 25.80  & 47.36  & 25.55    & 52.67    & 35.36 \\
			
			{} & {FWSVD~\cite{hsu2022languagemodel}}      & 8,173.21           & 8,615.71          & 8,024.67              &24.83   &25.84   &48.70   &25.64      &52.83    &35.57  \\
			
			{} & {ASVD~\cite{yuan2024asvd}}      & 6,977.57           & 15,539.44          & 4,785.15              &24.52   &25.13   &49.17   &25.48     &52.94    &35.45  \\
			
			{} & {SVD-LLM~\cite{wang2024svdllm}}      & 27.19           & 1,772.91        & {129.66}              & 23.44  & 31.65  & 51.14  & 28.38    & 54.57    & 37.83 \\ 
			
			\cmidrule{2-11} 
			{}                     & \cellcolor{purple!10}{\textbf{$\ours$}}  &\cellcolor{purple!10}\textbf{25.58}~\footnotesize($\downarrow$6\%)   & \cellcolor{purple!10}\textbf{593.14}~\footnotesize($\downarrow$67\%) & \cellcolor{purple!10}{\textbf{113.84}}~\footnotesize($\downarrow$12\%)   & \cellcolor{purple!10}23.24               & \cellcolor{purple!10}\textbf{34.18}               & \cellcolor{purple!10}\textbf{54.06}               & \cellcolor{purple!10}\textbf{28.88}                        & \cellcolor{purple!10}\textbf{55.50}             
			& \cellcolor{purple!10}\textbf{39.17}    \\

			\midrule
			{\multirow{5}{*}{60\%}} & {SVD}      & 65,186.67           & 79,164.00          & {70,381.00} & 22.94  & 24.49  & \textbf{51.85}  & 25.40   & 53.16   & 35.57 \\
			
			{} & {FWSVD~\cite{hsu2022languagemodel}}      & 27,213.30           & 24,962.80          & 47,284.87              &\textbf{26.91}   &25.38   &48.46   &25.61     & 51.96    & 35.66 \\
			
			{} & {ASVD~\cite{yuan2024asvd}}      & 10,003.57           & 15,530.19          & 9,983.83              &26.89   &26.68   &48.86   &25.76      & 51.80   &36.00  \\
			
			{} & {SVD-LLM~\cite{wang2024svdllm}}      & 89.90           & 2,052.89         & {561.00}              & 22.88  & 26.73  & 47.43  & 26.89    & \textbf{53.48}    & 35.48 \\ 
			\cmidrule{2-11} 
			{}                     & \cellcolor{purple!10}{\textbf{$\ours$}}  & \cellcolor{purple!10}\textbf{50.33}~\footnotesize($\downarrow$44\%)   & \cellcolor{purple!10}\textbf{1,216.95}~\footnotesize($\downarrow$41\%) & \cellcolor{purple!10}\textbf{239.18}~\footnotesize($\downarrow$57\%)  & \cellcolor{purple!10}24.69              & 
			\cellcolor{purple!10}\textbf{28.20}               & 
			\cellcolor{purple!10}51.22               & 
			\cellcolor{purple!10}\textbf{27.36}               & \cellcolor{purple!10}52.83                            & \cellcolor{purple!10}\textbf{36.87}   \\ 
			\bottomrule

		\end{tabular}
	}
	\label{tab:dataset_acc}
	\vspace{-2mm}
	\caption{Zero-shot performance comparison of LLaMA2-7B between $\ours$ and previous SVD compressed methods under 40\% to 60\% compression ratios. Evaluation on three language modeling datasets (measured by perplexity  ({$\downarrow$})) and five common sense reasoning datasets (measured by both individual and average accuracy ({$\uparrow$})) demonstrate the effectiveness of $\ours$.
	}
	\vspace{-4mm}
\end{table*}

\subsection{Adaptive SVD Compression Ratio}
\label{sec:adacr}

Previous studies on SVD compression typically apply a uniform compression ratio across all transformer layers of LLMs, overlooking the varying importance of different layers.
Inspired by~\citet{men2024shortgpt} and~\citet{dumitru2024change}, we propose \textbf{adaCR}, which adaptively determines the SVD compression ratio for each transformer layer, considering each layer's distinct impact on activations.

The importance of $\mathcal{W}$ can be measured by its impact on the input, which is quantified as the similarity between the input $\mathcal{X}$ and the output $\mathcal{Y}$ after passing through $\mathcal{W}$.
\begin{align}
	\mathcal{Y} &= \mathcal{WX}, \\
	\mathcal{I}(\mathcal{W}) &= \text{similarity}(\mathcal{X,Y}),
\end{align}
where $\mathcal{I}(\mathcal{W})$ denotes the layer-wise importance of $\mathcal{W}$. The similarity metric used can vary, and for simplicity, we adopt cosine similarity in our method.

Then, we normalize $\mathcal{I}(\mathcal{W})$ through mean centering to obtain the relative importance of $\mathcal{W}$:
\begin{align}
	\mathcal{I}_n(\mathcal{W}) = \mathcal{I}(\mathcal{W}) / \text{mean}(\mathcal{I}(\mathcal{W})).
\end{align}
After mean normalization, the average importance is 1. A value of $\mathcal{I}_n(\mathcal{W})$ greater than 1 indicates greater importance, while a value lower than 1 indicates lesser importance. The compression ratio of each layer will be adaptively adjusted based on the relative importance:
\begin{align}
	\mathcal{CR}(\mathcal{W}) = mrr + \mathcal{I}_n(\mathcal{W}) \cdot (trr - mrr),
\end{align}
where $mrr$ and $trr$ are the minimum and target retention ratios, respectively. Notably, $\mathcal{CR}(\mathcal{W}) = mrr$ when $\mathcal{I}_n(\mathcal{W})=0$, and $\mathcal{CR}(\mathcal{W}) = trr$ when $\mathcal{I}_n(\mathcal{W}) = 1$.

Given the compression ratio for the $i$-th layer by \textbf{adaCR}, we truncate the vectors of least singular values from both $\mathcal{U}_k^\sigma$ and ${\mathcal{V}_k^\sigma}^\top$ so that 
\begin{align}
	\mathcal{CR}(\mathcal{W}_i)=\frac{\#\text{params of }\mathcal{U}_k^\sigma + \#\text{params of }{\mathcal{V}_k^\sigma}^\top}{\#\text{params of }\mathcal{W}_i}.
\end{align}
As shown in~\cref{fig:layer_importance}, the importance of different layers varies. It can be observed that the first layer always weighs the most importance, suggesting that we should retain more weight on it. For the Llama family, the relative importance curve approximates a bowl shape, highlighting the significance of both the initial and final layers.

\section{Experiments}
\subsection{Setup}

We compare our $\ours$ with four baselines,  including vanilla SVD and SOTA SVD-based LLM compression methods FWSVD~\cite{hsu2022languagemodel}, ASVD~\cite{yuan2024asvd}, and SVD-LLM~\cite{wang2024svdllm}.

\begin{figure*}[htbp]
	\centering
	\includegraphics[width=1\textwidth]{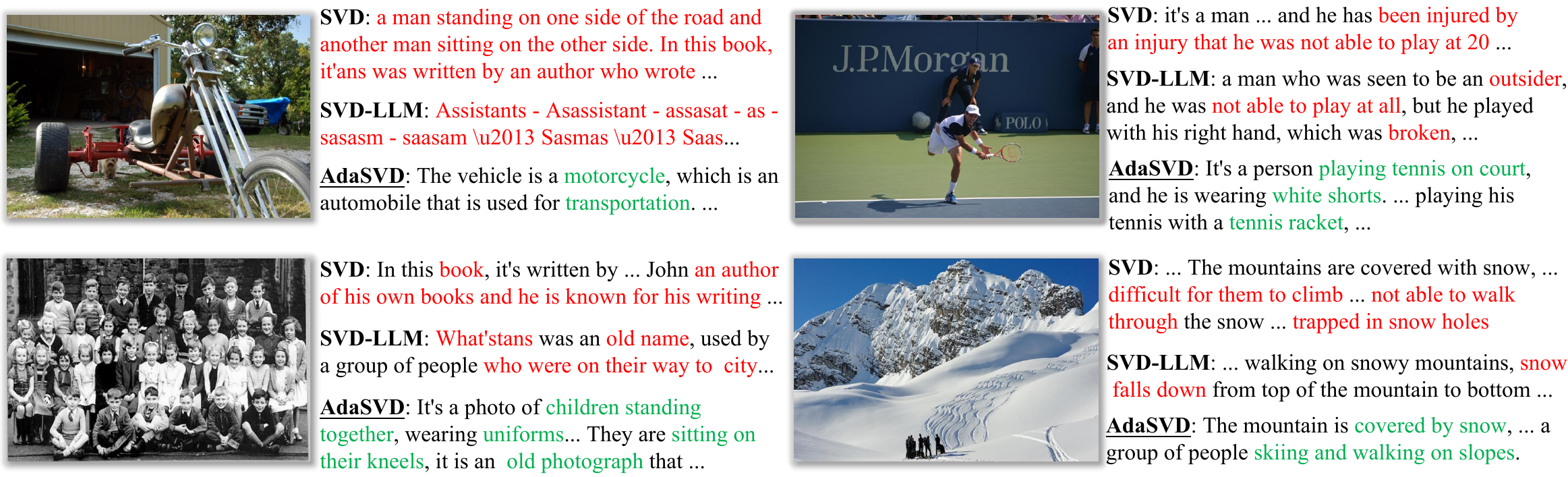}
	\vspace{-6mm}
	\caption{We perform image captioning by applying SVD, SVD-LLM~\cite{wang2024svdllm}, and our $\ours$ to LLaVA-7B model on the COCO dataset respectively, highlighting the \textcolor{color4}{correct} captions and \textcolor{red}{wrong} captions in different colors.}
	\vspace{-4.5mm}
	\label{fig:llava}
\end{figure*}

\begin{table}[t]
	\centering
	\vspace{1.5mm}
	\resizebox{\linewidth}{!}{%
		\begin{tabular}{c|cccc}
			\toprule
			\textsc{Method} &OPT-6.7B   & LLaMA2-7B    & Mistral-7B        & Vicuna-7B \\ \midrule
			SVD       & 18,607.24          & 65,186.67         & 30,378.35            & 78,704.50         \\
			FWSVD~\cite{hsu2022languagemodel}     & 8,569.56      & 27,213.30          & 5,481.24              & 8,185.66       \\
			ASVD~\cite{yuan2024asvd}      & 10,326.48         & 10,003.57         & 22,705.51             & 20,241.17          \\
			SVD-LLM~\cite{wang2024svdllm}      &   92.10      &   89.90       &  72.17            &  64.06        \\
			\midrule
			\rowcolor{purple!10}\textbf{$\ours$}  & \textbf{86.64}~\footnotesize($\downarrow$6\%)         & \textbf{50.33}~\footnotesize($\downarrow$44\%)           &  \textbf{67.22}~\footnotesize($\downarrow$7\%)          & \textbf{56.97}~\footnotesize($\downarrow$11\%)         \\ \bottomrule
		\end{tabular}
	}
	\vspace{-2mm}
	\caption{Perplexity ($\downarrow$) of four different LLMs -- OPT-6.7B, LLaMA 2-7B, Mistral-7B, and Vicuna-7B -- under 60\% compression ratio on WikiText-2, where $\ours$ shows consistent improvements. \label{tab:different_llm_acc}}
	\vspace{-4mm}
\end{table}

\noindent\textbf{Models and Datasets.$\quad$} 
To demonstrate the generalizability of our method, we evaluate the performance of $\ours$ and the baselines on four models from three different LLM families, including LLaMA2-7B~\cite{touvron2023llama2}, OPT-6.7B~\cite{zhang2022opt}, Mistral-7B~\cite{jiang2023mistral}, and Vicuna-7B~\cite{chiang2023vicuna}. We benchmark on eight datasets, including three language modeling datasets (WikiText-2~\cite{merity2016pointersentinelmixturemodels}, PTB~\cite{marcus1993building}, and C4~\cite{raffel2023exploring}) and five common-sense reasoning datasets (WinoGrande~\cite{sakaguchi2019winograndeadversarialwinogradschema}, HellaSwag~\cite{zellers2019hellaswagmachinereallyfinish}, PIQA~\cite{bisk2019piqareasoningphysicalcommonsense}, ARC-e~\cite{clark2018thinksolvedquestionanswering}, and Mmlu~\cite{hendryckstest2021}).
We use the LM-Evaluation-Harness framework~\cite{eval-harness} to evaluate the model performance on these zero-shot Question-Answering (QA) datasets.

\noindent\textbf{Implementation Details.$\quad$} 
To ensure a fair comparison, we followed ASVD~\cite{yuan2024asvd} and SVD-LLM~\cite{wang2024svdllm} to randomly select 256 samples from WikiText-2 as the calibration data and conduct data whitening before SVD truncation. 
All the experiments are conducted with PyTorch~\cite{paszke2019pytorch} and Huggingface~\cite{paszke1912imperative} on a single NVIDIA A100-80GB GPU.

\subsection{Main Results}

We evaluate the overall performance of $\ours$ from three aspects: \textbf{(1)} performance under different compression ratios \textbf{(40\%, 50\%, 60\%, 70\%, and 80\%)}, \textbf{(2)} performance on different LLMs. \textbf{(3)} performance on visual language models. Some performance evaluation results and generated contents by the compressed LLMs are included in the supplementary file to provide a more straightforward comparison.

\noindent\textbf{Performance under Different Compression Ratios.$\quad$}
First, we evaluate the performance of LLaMA2-7B compressed by $\ours$, vanilla SVD and the SOTA method SVD-LLM~\cite{wang2024svdllm} under compression ratios ranging from 40\% to 80\% on all $8$ datasets, as shown in~\cref{tab:dataset_acc}. 
On the three language modeling datasets, $\ours$ consistently outperforms vanilla SVD, and SVD-LLM across all the compression ratios. 
More importantly, $\ours$ exhibits significant advantages over the baselines under higher compression ratios. 
These results indicate that $\ours$ is more effective in compressing LLMs for more resource-constrained devices such as smartphones and IoT devices, which often have limited memory and processing capabilities.
On the five common sense reasoning datasets, 
$\ours$ also maintains its edge and performs better than the best-performing baseline on most of the datasets and consistently achieves higher average accuracy across all the compression ratios. Due to page limitations, comparisons for 70\% and 80\% compression ratios are provided in the supplementary file.

\noindent\textbf{Performance on Different LLMs.$\quad$}
To demonstrate the generability of $\ours$ across different LLMs, we compare $\ours$ and the baselines on four different models OPT-6.7B, LLaMA2-7B, Vicuna-7B, and Mistral-7B -- under 60\% compression ratio on WikiText-2.
As shown in~\cref{tab:different_llm_acc},  
$\ours$ consistently outperforms vanilla SVD, FWSVD, ASVD and SVD-LLM on all LLMs, and exhibits more stable performance across different LLMs, especially compared to vanilla SVD and FWSVD. We reproduce FWSVD, ASVD, and SVD-LLM using their official GitHub repositories. FWSVD and ASVD fail on these LLMs with compression ratios under 60\%, whereas SVD-LLM and $\ours$ maintain reasonable perplexity in such cases.

\noindent\textbf{Performance on Visual Language Models.$\quad$} Note that our $\ours$ can also be applied to visual language models (VLMs) like LLaVA~\cite{liu2023visual}. Following~\citet{lin2024awq}, we apply SVD compression to the language part of the VLMs since it dominates the model size. As shown in~\cref{fig:llava}, $\ours$ shows better image captioning results than vanilla SVD and SVD-LLM on COCO dataset~\cite{chen2015microsoft} under 40\% compression ratio.
More image captioning comparisons with various compression ratios can be found in supplementary file.

\subsection{Ablation Study}

We provide extensive ablation study results in ~\cref{tab:ablations} to show the effect of some key components in our work. 

\begin{table*}[t]
	\vspace{-2mm}

	\subfloat[\small Effectiveness of Adaptive Compensation \label{tab:adacomp}]{
		\scalebox{0.8}{\begin{tabular}{l@{\hskip 9pt}c@{\hskip 12pt}c@{\hskip 12pt}c@{\hskip 12pt}c@{\hskip 12pt}c}
				\toprule
				\rowcolor{color3}
				\textbf{Method} & \textbf{Tgt. CR} & \textbf{adaComp}  &\textbf{WikiText2 $\downarrow$} & \textbf{PTB $\downarrow$} & \textbf{C4 $\downarrow$}   \\
				\midrule
				SVD-LLM & 40\% & \ding{55} &16.11  &719.44   &61.95 \\
				\cdashline{1-6} \addlinespace[0.2em]
				$\ours$ & 40\% & \ding{55} & 15.47  &406.83  &66.29  \\
				\rowcolor{purple!10}$\ours$ & 40\% &  \ding{51} &14.76  &304.62  &56.98  \\
				\midrule
				SVD-LLM & 50\% & \ding{55} & 27.19&1,772.91 &129.66 \\
				\cdashline{1-6} \addlinespace[0.2em]
				$\ours$ & 50\% & \ding{55} & 30.00 & 1101.15 &166.02  \\
				\rowcolor{purple!10}$\ours$ & 50\% &  \ding{51} &25.58  & 593.14 &113.84  \\
				\midrule
				SVD-LLM & 60\% & \ding{55} &89.90 &2,052.89 &561.00 \\
				\cdashline{1-6} \addlinespace[0.2em]
				$\ours$ & 60\% & \ding{55} &78.82  &6,929.39  &339.31  \\
				\rowcolor{purple!10}$\ours$ & 60\% &  \ding{51} &50.33  &1,216.95 &239.18  \\
				\bottomrule
	\end{tabular}}}\hfill
	\subfloat[\small Effectiveness of Adaptive Compression Ratio\label{tab:adacr}]{ 
		\scalebox{0.8}{
			\begin{tabular}{l@{\hskip 9pt}c@{\hskip 12pt}c@{\hskip 12pt}c@{\hskip 12pt}c@{\hskip 12pt}c}
				\toprule
				\rowcolor{color3}\textbf{Method} &\textbf{Tgt. CR} &\textbf{CR} & \textbf{WikiText2 $\downarrow$} & \textbf{PTB $\downarrow$} & \textbf{C4 $\downarrow$} \\
				\midrule
				SVD-LLM & 40\% & Const &16.11  &719.44   &61.95  \\
				\cdashline{1-6} \addlinespace[0.2em]
				$\ours$ & 40\% & Const & 15.38  &617.11  &60.43  \\
				\rowcolor{purple!10}$\ours$ & 40\% & Adapt &14.76  &304.62  &56.98  \\
				\midrule
				SVD-LLM & 50\% & Const & 27.19&1,772.91 &129.66 \\
				\cdashline{1-6} \addlinespace[0.2em]
				$\ours$ & 50\% & Const & 27.33 & 1,177.53 &126.85  \\
				\rowcolor{purple!10}$\ours$ & 50\% & Adapt &25.58  & 593.14 &113.84  \\
				\midrule
				SVD-LLM & 60\% & Const &89.90 &2,052.89 &561.00 \\
				\cdashline{1-6} \addlinespace[0.2em]
				$\ours$ & 60\% & Const &69.46  &2,670.20  &336.90 \\
				\rowcolor{purple!10}$\ours$ & 60\% &  Adapt &50.33  &1,216.95 &239.18   \\
				
				\bottomrule
	\end{tabular}}} \\
	\subfloat[\small Iteration Number for Adaptive Compression \label{tab:num_iter}]{
		\scalebox{0.8}{\begin{tabular}{l@{\hskip 9pt}c@{\hskip 12pt}c@{\hskip 12pt}c@{\hskip 12pt}c@{\hskip 12pt}c}
				\toprule
				\rowcolor{color3}
				\textbf{Method} &\textbf{Tgt. CR} & \textbf{\#Iteration} &\textbf{WikiText2 $\downarrow$} &\textbf{PTB $\downarrow$} &\textbf{C4 $\downarrow$}   \\
				\midrule
				SVD-LLM & 40\% & - &16.11  &719.44   &61.95  \\
				\cdashline{1-6} \addlinespace[0.2em]
				\rowcolor{purple!10}$\ours$ & 40\% & 1 &14.76  &304.62  &56.98   \\
				$\ours$ &  40\% &  3 & 15.47  &249.41   &57.28   \\
				$\ours$ & 40\% & 15  & 15.84 &257.96  & 57.39 \\
				\midrule
				SVD-LLM & 50\% & - & 27.19&1,772.91 &129.66  \\
				\cdashline{1-6} \addlinespace[0.2em]
				\rowcolor{purple!10}$\ours$ & 50\% & 1 &25.58  & 593.14 &113.84  \\
				$\ours$ & 50\% & 3  &27.11  &844.09 &115.51\\
				$\ours$ & 50\% & 15 &27.45  & 812.21 & 110.35  \\
				\midrule
				SVD-LLM & 60\% & - &89.90 &2,052.89 &561.00  \\
				\cdashline{1-6} \addlinespace[0.2em]
				\rowcolor{purple!10}$\ours$ & 60\% & 1 &50.33  &1,216.95 &239.18  \\
				$\ours$ & 60\% & 3 &64.12  &3,546.45 &301.19  \\
				$\ours$ & 60\% & 15 &62.34  &4,293.79 &267.29  \\
				\bottomrule
	\end{tabular}}}\hfill
	\subfloat[\small Minimum Retention Ratio for Adaptive CR\label{tab:mrr}]{ 
		\scalebox{0.8}{\begin{tabular}{l@{\hskip 9pt}c@{\hskip 12pt}c@{\hskip 12pt}c@{\hskip 12pt}c@{\hskip 12pt}c}
				\toprule
				\rowcolor{color3}
				\textbf{Method} &\textbf{Tgt. CR} & \textbf{MRR} &\textbf{WikiText2 $\downarrow$} &\textbf{PTB $\downarrow$} &\textbf{C4 $\downarrow$}   \\
				\midrule
				SVD-LLM & 40\% & - &16.11  &719.44   &61.95  \\
				\cdashline{1-6} \addlinespace[0.2em]
				$\ours$ & 40\% & 0.40 &15.01  &223.19 &57.17  \\
				$\ours$ & 40\% & 0.45 &14.85  &241.90  &57.08   \\
				\rowcolor{purple!10}$\ours$ & 40\% & 0.50 &14.76  &304.62 &56.98  \\
				\midrule
				SVD-LLM & 50\% & - & 27.19&1,772.91 &129.66  \\
				\cdashline{1-6} \addlinespace[0.2em]
				\rowcolor{purple!10}$\ours$ & 50\% & 0.40 &25.58  &593.14 &113.84  \\
				$\ours$ & 50\% & 0.45 &26.01  & 814.63  &117.58  \\
				$\ours$ & 50\% & 0.50 &27.33  &1,177.53 &126.85  \\
				\midrule
				SVD-LLM & 60\% & - &89.90 &2,052.89 &561.00  \\
				\cdashline{1-6} \addlinespace[0.2em]
				\rowcolor{purple!10}$\ours$ & 60\% & 0.30 & 50.33 &1,216.95 &239.18  \\
				$\ours$ & 60\% & 0.35 & 53.17 &1,608.19 &256.66  \\
				$\ours$ & 60\%  & 0.40 & 60.08 &2,137.29 &294.26 \\
				\bottomrule
	\end{tabular}}}
	\vspace{-2mm}
	\caption{Ablation studies on LLaMA-2-7B. Results are measured by perplexity, with best results highlighted in \colorbox{purple!10}{\phantom{0000}}.\label{tab:ablations}}
	\vspace{-3mm}
\end{table*}

\noindent\textbf{Effectiveness of Adaptive Compensation. $\quad$} To validate the effectiveness of the proposed \textbf{adaComp}, we compare the PPL results of Llama2-7B with and without \textbf{adaComp} on Wikitest-2, PTB, and C4 datasets in~\cref{tab:adacomp}. Results of 70\% and 80\% compression ratios can be found in the supplementary file.
It can be observed that $\ours$ consistently outperforms SVD-LLM after applying \textbf{adaComp}, and the performance gap is more significant under high compression ratios (\textit{i.e.}, 60\%, 70\%, and 80\%).

\noindent\textbf{Iteration Number. $\quad$} 
To investigate the impact of the number of \textbf{adaComp} iterations under different compression ratios, we perform an ablation study with 1, 3, and 15 iterations, as shown in~\cref{tab:num_iter}. Results for 70\% and 80\% compression ratios are provided in the supplementary file. At lower compression ratios (\textit{e.g.}, 40\%, 50\%, and 60\%), it is observed that just 1 iteration of \textbf{adaComp} already outperforms the state-of-the-art method, SVD-LLM. However, increasing the number of iterations may lead to overfitting due to the limited calibration data, resulting in a performance drop. In contrast, at higher compression ratios (\textit{e.g.}, 70\% and 80\%), additional iterations lead to performance improvements, indicating that $\ours$ is more effective in high compression ratio scenarios where previous methods still struggle. This highlights the importance of balancing the number of iterations with the available data to avoid over-optimization, especially in low compression scales.

\begin{table}[t]
	\centering
	
	\resizebox{1\linewidth}{!}{
	\begin{tabular}{c|c|c|c|c|c}
		\toprule
		{\textsc{Ratio}}       & {\textsc{Method}}   & {\textsc{GPTQ-INT4}} & WikiText-2{$\downarrow$} & PTB{$\downarrow$} & {C4{$\downarrow$}}  \\ \midrule
		{\color[HTML]{9B9B9B}0\%}  & {\color[HTML]{9B9B9B}Original} & \ding{55}  & {\color[HTML]{9B9B9B}5.68}   & {\color[HTML]{9B9B9B}8.35}     & {\color[HTML]{9B9B9B}7.34}    \\ \midrule
		{\multirow{4}{*}{40\%}} & {SVD-LLM}  & \ding{55}     &16.11  &719.44            & {61.95}               \\
		& {SVD-LLM}  & \ding{51}     & 33.56          & 1,887.50          & {184.61}               \\
		& \cellcolor{purple!10}{\textbf{$\ours$}} & \cellcolor{purple!10}\ding{55}  & \cellcolor{purple!10}\textbf{14.76}   & \cellcolor{purple!10}\textbf{304.62} & \cellcolor{purple!10}{\textbf{56.98} }      \\ 
		& \cellcolor{purple!10}{\textbf{$\ours$}} & \cellcolor{purple!10}\ding{51}  & \cellcolor{purple!10}\textbf{22.55}   & \cellcolor{purple!10}\textbf{844.21} & \cellcolor{purple!10}{\textbf{106.41} }      \\
		\midrule
		
		{\multirow{4}{*}{50\%}} & {SVD-LLM}  & \ding{55}     & 27.19           & 1,772.91  & {129.66}                     \\ 
		& {SVD-LLM}  & \ding{51}     & 41.70           & 2,335.65             & {291.62}         \\ 
		
		& \cellcolor{purple!10}{\textbf{$\ours$}} & \cellcolor{purple!10}\ding{55}  &\cellcolor{purple!10}\textbf{25.58}   & \cellcolor{purple!10}\textbf{593.14}    & \cellcolor{purple!10}{\textbf{113.84} }  \\ 
		& \cellcolor{purple!10}{\textbf{$\ours$}} & \cellcolor{purple!10}\ding{51}  &\cellcolor{purple!10}\textbf{37.34}   & \cellcolor{purple!10}\textbf{1,326.55}    & \cellcolor{purple!10}{\textbf{203.11} }  \\

		\midrule
		
		{\multirow{4}{*}{60\%}} & {SVD-LLM}  & \ding{55}     & 89.90           & 2,052.89          & {561.00}              \\ 
		& {SVD-LLM}  & \ding{51}     & 119.46           & 3,136.60              & {723.80}         \\ 
		& \cellcolor{purple!10}{\textbf{$\ours$}} & \cellcolor{purple!10}\ding{55}  & \cellcolor{purple!10}\textbf{60.08}   & \cellcolor{purple!10}\textbf{2,137.28}   & \cellcolor{purple!10}{\textbf{294.26} }    \\ 
		& \cellcolor{purple!10}{\textbf{$\ours$}} & \cellcolor{purple!10}\ding{51}  & \cellcolor{purple!10}\textbf{82.08}   & \cellcolor{purple!10}\textbf{1,705.19} & \cellcolor{purple!10}{\textbf{379.96} }     \\

		\midrule
		
		{\multirow{4}{*}{70\%}} & {SVD-LLM}   & \ding{55}    & 125.16          & 6,139.78        & {677.38}                \\ 
		& {SVD-LLM}   & \ding{51}    & 159.53         & 2,115.44           & {848.24}            \\ 
		& \cellcolor{purple!10}{\textbf{$\ours$}} & \cellcolor{purple!10}\ding{55}  & \cellcolor{purple!10}\textbf{107.90}   & \cellcolor{purple!10}\textbf{5,027.62}  & \cellcolor{purple!10}{\textbf{441.33} }  
		\\  
		& \cellcolor{purple!10}{\textbf{$\ours$}} & \cellcolor{purple!10}\ding{51}  & \cellcolor{purple!10}\textbf{118.75}   & \cellcolor{purple!10}\textbf{1,606.94}   & \cellcolor{purple!10}{\textbf{466.64} }  
		\\

		\midrule
		
		{\multirow{4}{*}{80\%}} & {SVD-LLM}   & \ding{55}    & 372.48           & 6,268.53     & {1,688.78}                 \\ 
		& {SVD-LLM}  & \ding{51}     & 420.25           & 3,716.08 & {1,996.42}                     \\ 
		& \cellcolor{purple!10}{\textbf{$\ours$}} & \cellcolor{purple!10}\ding{55}  & \cellcolor{purple!10}\textbf{206.51}   & \cellcolor{purple!10}\textbf{6,613.44}   & \cellcolor{purple!10}{\textbf{679.66} }   \\ 
		& \cellcolor{purple!10}{\textbf{$\ours$}} & \cellcolor{purple!10}\ding{51}  & \cellcolor{purple!10}\textbf{214.51}   & \cellcolor{purple!10}\textbf{2,728.78} & \cellcolor{purple!10}{\textbf{654.79} }   \\  \bottomrule

	\end{tabular}
}
\vspace{-2mm}
\caption{$\ours$ with weight quantization method GPTQ.
}
\vspace{-5mm}
\label{tab:svd+quant}
\end{table}

\noindent\textbf{Effectiveness of Adaptive Compression Ratio. $\quad$} 
To validate the effectiveness of our \textbf{adaCR}, we compared the results after removing \textbf{adaCR} (\textit{i.e.}, using constant compression ratios for all layers) from $\ours$. As shown in~\cref{tab:adacr}, $\ours$ already outperforms SOTA SVD-LLM without using \textbf{adaCR}, while integrating \textbf{adaCR} can further enhance the performance across all compression ratios.

\noindent\textbf{Minimum Retention Ratio. $\quad$} The minimum retention ratio ($mrr$) in \textbf{adaCR} is also crucial, and we investigate the impact of different $mrr$ values in~\cref{tab:mrr} for 40\%, 50\%, and 60\% compression ratios (70\% and 80\% in supplementary file). It can be observed that $mrr$ remains relatively robust at lower compression ratios (40\% and 50\%), while contributing more at higher compression ratios (60\%).


\subsection{Integrate with Weight Quantization}
Similar to previous SVD-based compression methods~\cite{hsu2022fwsvd,yuan2024asvd,wang2024svdllm}, our $\ours$ is orthogonal to other types of compression techniques. Following~\citet{wang2024svdllm}, we integrate $\ours$ with the widely used weight quantization method GPTQ~\cite{frantar2022gptq}. As shown in~\cref{tab:svd+quant}, we compare $\ours$ with SVD-LLM~\cite{wang2024svdllm} on the LLaMA2-7B model, using different compression ratios (40\%, 50\%, 60\%, 70\%, and 80\%) across the WikiText-2, PTB, and C4 datasets. The results demonstrate that, when combined with the 4-bit weight quantization method GPTQ, $\ours$ also consistently outperforms SOTA SVD-LLM across all compression ratios. Under high compression ratios (\ie, 60\%, 70\%, and 80\%), $\ours$ + GPTQ-INT4 even surpasses SVD-LLM.

\vspace{-2mm}
\section{Conclusion}
\vspace{-1.5mm}
In this work, we propose $\ours$, an adaptive SVD-based compression method for LLMs. $\ours$ 
first proposes \textbf{adaComp}, which adaptively compensates for the error caused by the truncation of singular matrices, efficiently reducing compression error without requiring additional training. Furthermore, $\ours$ proposes \textbf{adaCR}, which adaptively assigns compression ratios based on the importance of each layer, further enhancing performance while maintaining the same target compression rate. Both strategies effectively minimize SVD compression errors, particularly at high compression ratios. Our experiments on multiple open-source LLM and VLM families demonstrate that $\ours$ pushes the performance boundary beyond the current state-of-the-art SVD-based LLM compression methods.

{
\small
\bibliographystyle{ieeenat_fullname}
\bibliography{main}
}


\end{document}